\documentclass{article}
\usepackage{spconf,amsmath,graphicx}

\title{Super-realtime facial landmark detection and shape fitting by deep regression of shape model parameters}
\name{Marcin Kopaczka, Justus Schock, Dorit Merhof}
\address{Institute of Imaging and Computer Vision, RWTH Aachen University, Germany}
\begin{document}
%
\maketitle
\begin{abstract}
We present a method for highly efficient landmark detection that combines deep convolutional neural networks with well established model-based fitting algorithms. Motivated by established model-based fitting methods such as active shapes, we use a PCA of the landmark positions to allow generative modeling of facial landmarks. Instead of computing the model parameters using iterative optimization, the PCA is included in a deep neural network using a novel layer type. The network predicts model parameters in a single forward pass, thereby allowing facial landmark detection at several hundreds of frames per second. Our architecture allows direct end-to-end training of a model-based landmark detection method and shows that deep neural networks can be used to reliably predict model parameters directly without the need for an iterative optimization. The method is evaluated on different datasets for facial landmark detection and medical image segmentation. PyTorch code is freely available at https://github.com/justusschock/shapenet
\end{abstract}
\section{Introduction and previous work}
\label{sec:intro}

Landmark detection for faces and anatomical structures has been a key research area in computer vision for decades. Before current deep learning methods otuperformed them, optimzier-driven generative approaches were among the best performing landmark detection methods. Well-known examples include active shape models (ASMs)~\cite{cootes1995active}, active appearance models (AAMs)~\cite{cootes2001active}~\cite{matthews2004active} and constrained local models (CLMs)~\cite{cristinacce2006feature}. While they vary strongly in detail, their common root is a parametrized statistical landmark shape model that allows landmark fitting by adjusting the parameters of the model, thereby generating a shape instance that is updated by iteratively optimizing the parameter values until optimal landmark positions are found. While being precise when trained with a sufficient amount of training data, these methods have recently been surpassed by machine learning-based landmark detection algorithms. Notable current methods include deep alignment networks~\cite{kowalski2017deep}, stacked hourglass networks~\cite{yang2017stacked} and the HyperFace framework~\cite{ranjan2019hyperface}, which allows not only landmark detection, but provides a full end-to-end solution for face and landmark detection and subsequent analysis tasks. All of these methods outperform the aforementioned classical approaches in terms of fitting accuracy and robustness. However, only a small number of both learning-based and conventional methods are able to achieve real-time performance on a 30 fps video while still providing precise detections. Even less methods are able to perform detection at significantly higher speeds, thereby leaving room for additional real-time analysis on top of the detection. The most notable approaches in high-speed landmark detection are the regression-based methods published in~\cite{ren2014face} and~\cite{kazemi2014one} that provide landmark detection in the order of a milisecond. These algorithms also apply a statistical point distribution model, however a regression of the optimal model parameters in a single pass is performed instead of performing time-consuming iterative optimization to find the optimal model parameters.

\begin{figure}[t]
	\begin{minipage}[b]{1.0\linewidth}
	\centering
	\centerline{\includegraphics[width=8.5cm]{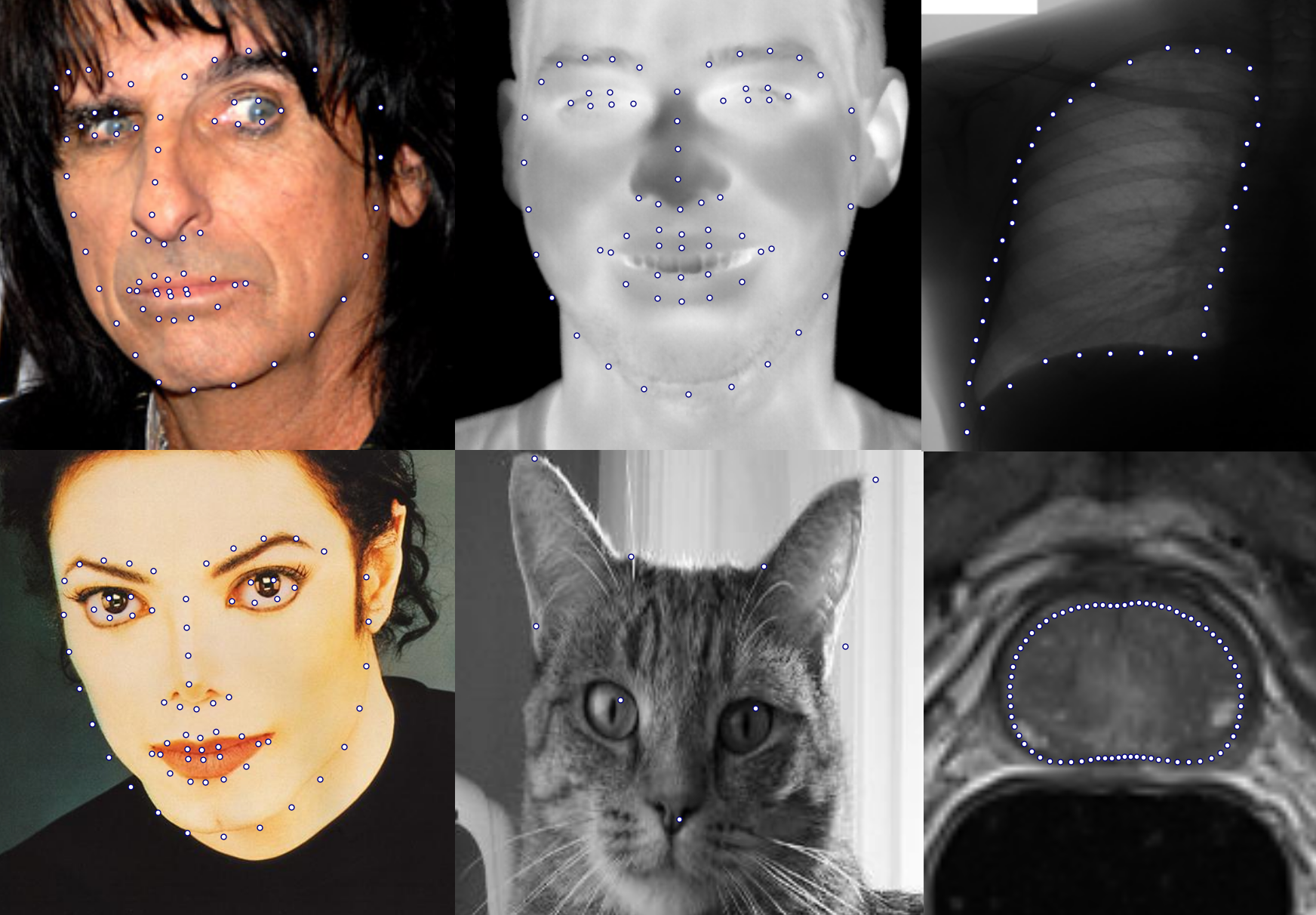}}
\end{minipage}
	\caption{Median results of landmarks localized with our method on different publicly available datasets. Left column: 300W outdoor (top) and indoor (bottom) datasets. Center column: Thermal IR Database (top), cat facial landmark dataset (bottom). Right column: JSRT Lung radiograph database (top), PROMISE prostate dataset (bottom). The predictions in all cases were made at 410 fps on a Geforce RTX 2080 Ti.}
\end{figure}


In our work, we also apply a regression  and perform the prediction of parameters of a statistical shape model using a deep neural network. This approach allows combining the speed of regressive methods with the robustness of neural networks and the benefit of obtaining plausible facial shapes as the result of a shape-constrained generative modeling algorithm. A conceptionally similar approach of predicting PCA parameters with a neural network has been published for medical data in~\cite{Bhalodia_2018}, however the network presented there did not contain an instance of the shape model, requiring explicit computation of the model instance outside of the network. In our proposed method, the PCA-based landmark localization is an inherent part of the network, thereby allowing efficient end-to-end training with a fully differentiable loss function and a drastically higher prediction speed. In~\cite{Milletari2017IntegratingSP}, the authors already included the PCA into the network to segment organs in medical images, however the network structure presented there was tailored to analyze spatiotemporal medical data and included several additional domain-specific steps to allow precise fitting in ultrasound images.
In our presented work, we instead capsule the point distribution model into a dedicated network layer, thereby allowing using it with a number of existing architectures with minimal effort. Additionally, we introduce a lightweight fully convolutional feature extraction network that allows extremly fast deep facial landmark detection. Finally, we include global translation, scaling and rotation parameters to remove the effect of global transformations on the landmark coordinates, allowing using a model that predicts local displacements exclusively.

\section{Methods}
\label{sec:methods}

In this section, we describe our proposed network topology and the training procedure in detail.

\subsection{Feature extraction}
\label{sec:Features}

The network consists of a feature extraction stage for parameter regression and a PCA combination layer where the computed shape eigenvectors are multiplied with the predicted parameters and form the final landmark positions. Our suggested feature extraction stage receives an image with the face as an input and its regressive output consists of the predicted weights of the PCA parameters as well as homogeneus global transform parameters. It is implemented as a set of convolutional layers with no fully connected layers to reduce the required number of network parameters. Additionally, we use spatially seperated convolutions along both image axes to further reduce the parameter count. Table~\ref{tab:Layers} shows a detailed layer overview.

\begin{figure}[t]
	\begin{minipage}[b]{1.0\linewidth}
		\centering
		\centerline{\includegraphics[width=8.5cm]{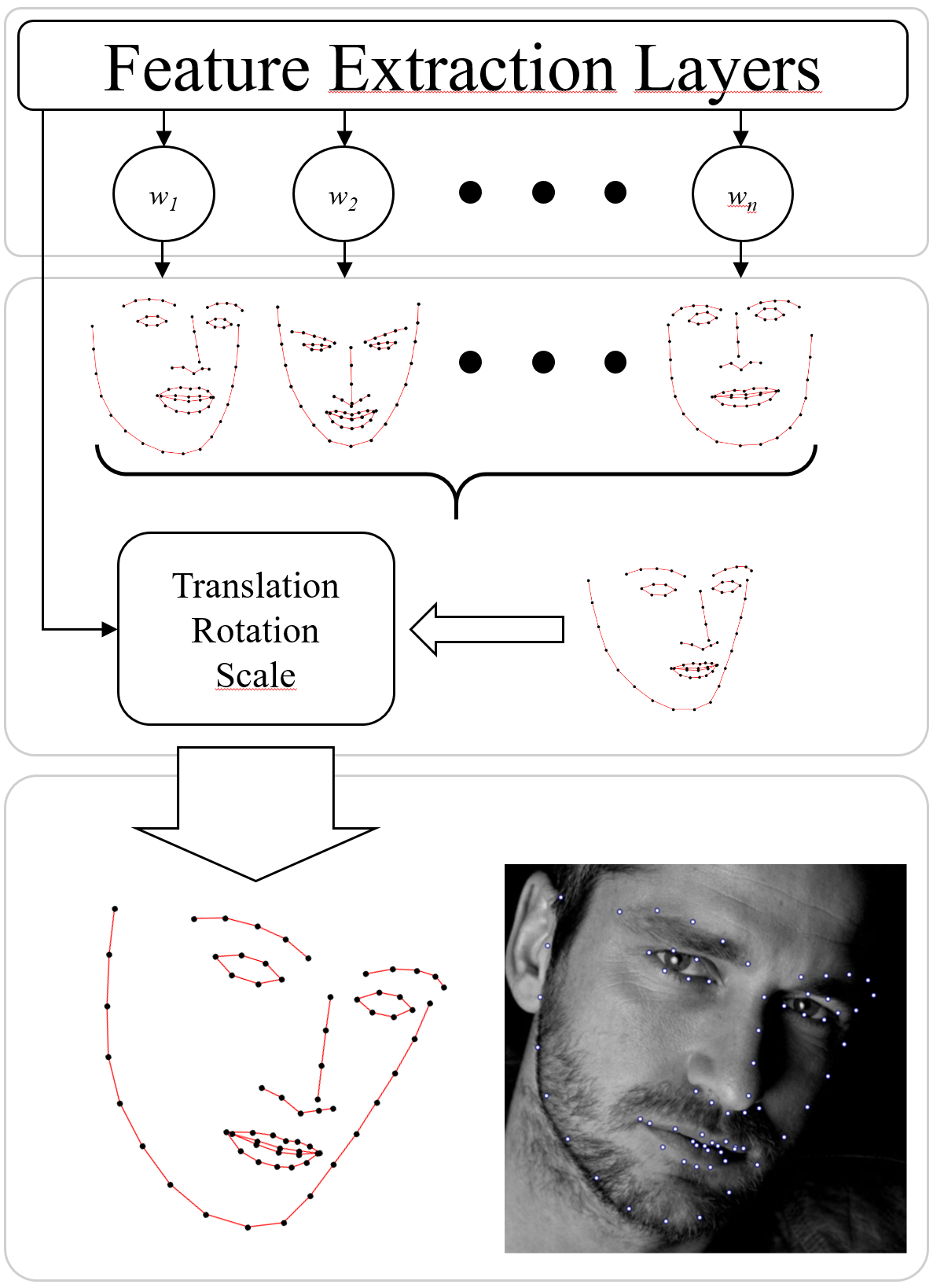}}
	\end{minipage}
	\caption{The PCA layer as part of the neural network. After regression of the model parameters by the feature extraction layers (top block), the shape in canonical coordinates is computed as linear combination of the weighted eigenvector shapes (center block). Subsequently, the global transform using the parameters comouted by the freature extraction stage is applied to transform the coordinates into the image reference frame (bottom block).}
	\label{Fig:PCALayer}
\end{figure}

\begin{table*}[]
	\centering
	\begin{tabular}{|l|l|l|l|l|l|l|l|l|l|}
		\hline
		& 1    & 2  & 3    & 4   & 5    & 6   & 7    & 8   & 9    \\ \hline
		Layer           & C2DB & DN & C2DB & DN  & C2DB & DN  & C2DB & DN  & C2DB \\ \hline
		Output channels & 64   & 64 & 128  & 128 & 256  & 256 & 512  & 256 & 128  \\ \hline
		Frequency       & 2    & 1  & 2    & 1   & 4    & 1   & 4    & 1   & 3    \\ \hline
	\end{tabular}
	\caption{Feature extraction network architecture. C2DB - Block of convolutional layers. Each layer is a 2D convolution with a (3 x 3) kernel at a stride of 1 and no zero padding, followed by a ReLu layer. A block consists of a concatenation of the same layer several times, with the number given in the 'Frequency' row. DN - Downsampling and normalization. A (3 x 3) 2D convolution with a stride of 2 followed by a ReLu layer and instance normalization.}
	\label{tab:Layers}
\end{table*}

\subsection{PCA layer with homogeneus transform}
\label{sec:PCALayer}

The design of the PCA layer is shown in Fig.\ref{Fig:PCALayer}.  
As input, the layer receives parameters for the global homogeneus transform (one scale parameter, one rotation parameter and two translation parameters) and one input for every PCA eigenvector. The eigenvectors are computed from the training data set in a preprocessing step and loaded into the network before training. The eigenvectors are normalized with their eigenvalues to ensure a consistent value range, thereby increasing convergence speed during training. They remain constant and are not changed during training by backpropagation unlike the remaining network parameters. Additionally, the mean shape computed by PCA is loaded into the network and receives a fixed weight of 1.

To compute landmark positions from the layer inputs, the eigenvectors are multiplied with their their corresponding weights and the results are linearly combined together with the mean shape to obtain a shape instance in local coordinates. Subsequently, the landmarks are converted into homogeneus coordinates and a transform matrix is computed from the regressed global transform parameters and applied to the local landmark positions. Seperating local displacements from global transformations allows using the PCA-based point distribution model to describe local shape variances exclusively, thereby increasing both the model's capability to include a larger number of relevant shape variations with less parameters as well as the layer's capability to predict facial landmarks with large global displacements. As a result, the layer returns the final landmark coordinates in the input image. The whole structure is fully differentiable, allowing including it at any stage of a deep neural network.

Note that the design of the PCA layer is generic, therefore it can be applied to perform regression of any data distribution that can be described by PCA.
We assume that this generic layer layout may be applied not only in image processing problems, but for any task in which an underlying point distribution representable as a statistic PCA-based model can be applied.

\subsection{Data Preparation and Network training}
\label{sec:training}

The network architecture is chosen as combination of a feature extraction network and the PCA layer.
Training requires a database such as HELEN~\cite{le2012interactive} or LFPW~\cite{belhumeur2013localizing} containing consistent landmark positions. The images are cropped around the landmarks and the crops are resized to uniform size (in our case, we have chosen 224 x 224 pixels as this is the default size of many CNN reference architectures). Subsequently, a PCA of the landmark coordinates of the training images is computed and the mean shape and a selected number of shape variations provided by the eigenvectors are stored in the PCA layer of the network. The number of included shape variations can be chosen arbitrarily as layer parameter. Since we explicitly remove rotational variaton from the PCA, the rotation of the training images is normalized by aligning the eye centers horizontally before applying PCA. Subsequently, rotation is introduced into the network by adding it during the augmentation stage.

The network input consists of the cropped images which are processed by the feature extraction and PCA layers. The entire network is trained end-to-end with a fast point distance loss such as L1 or MSE that aim at minimizing the distance between predicted and ground truth landmark positions.

\section{Experiments and Results}
\label{sec:experiments}
\begin{figure}[h]
	\begin{minipage}[b]{1.0\linewidth}
		\centering
		\centerline{\includegraphics[width=8.5cm]{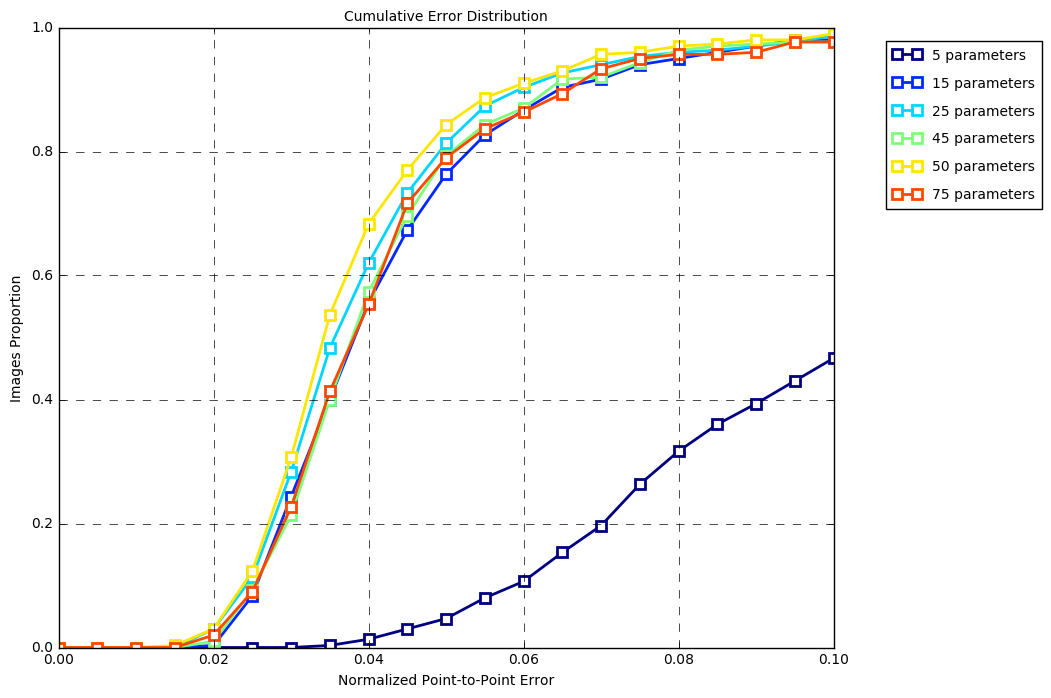}}
	\end{minipage}
	\caption{Effect of the number of shape parameters on the fitting precision.}
	\label{fig:numParams}
\end{figure}
To analyze an optimal number of shape parameters for facial landmark detection, we trained our method on a model to predict 5 to 75 PCA-generated shape parameters. The training sets of the HELEN and LFPW databases were combined and the network was evaluated with the 300W indoor set~\cite{sagonas2016300}. The results are shown in Fig.~\ref{fig:numParams}. It can be seen that while 5 parameters yield the lowest performance, adding more than 15 parameters does not change the fitting accuracy significantly or systematically in the case of facial landmark detection. Note that unlike iterative approaches where the number of parameters increases computation time, the runtime of our method is identical regardless of the number of parameters (410 fps on a Geforce RTX2080 Ti at half precision, this number applies to all datasets presented here since we used the same architecture regardless of database and imaging modality). The error metric used for all following evaluations is the normalized point-to-point error as defined in~\cite{zhu2012face}, where the root mean square error for each landmark is computed, averaged by the number of landmark points and normalized by the mean of the width and height of the ground truth landmarks. This method allows efficient comparison of landmark detection algorithms regardless of input image size and local displacements such as head pose.

\begin{figure}[h]
	\begin{minipage}[b]{1.0\linewidth}
		\centering
		\centerline{\includegraphics[width=8.5cm]{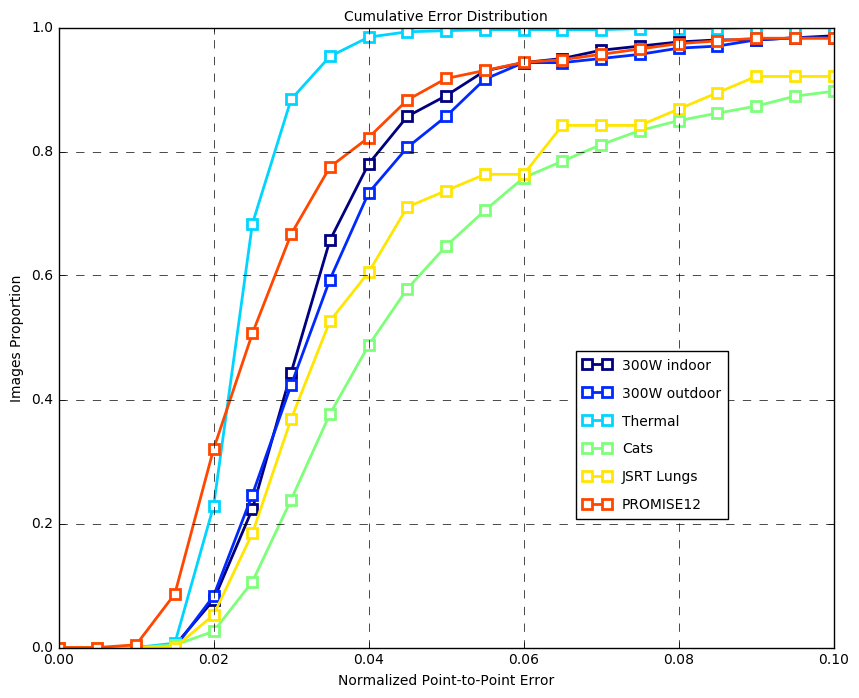}}
	\end{minipage}
	\caption{Normalized errors over the different evaluated datasets.}
	\label{fig:all_errors}
\end{figure}

To evaluate the applicability to other problems and imaging modalities, we analyzed the performance of the architecture on a number of other landmark localization tasks and different datasets:
\begin{itemize}
	\item The \textbf{300 faces in the wild (300W)} dataset ~\cite{sagonas2016300} containing 600 manually nnotated faces in an unconstrained environment (300 indoor and 300 outdoor images in two seperate sets). These images show real-world face data with a large variety of head poses, illumination conditions and facial expressions. Since 300W does not provide a training set, we used the LFPW and HELEN databases for training as they offer comparable image complexity. 
	\item The \textbf{thermal face database} presented in~\cite{8584115}, consisting of 2935 thermal face images manually annotated with the 68-point landmark scheme also used in our versions of LFPW, HELEN and 300W. The database is split into 2357 training and 578 test images. We trained the method with 25 shape parameters and 4 global transform parameters. The fitting precision is higher than for the visual 300W database, however the thermal dataset contains images acquired in a laboratory environment which are less challenging than the used RGB datasets which were acquired under unconstrained conditions and show higher variance regarding head pose, illumination conditions and image size.
	\item the \textbf{Kaggle cat dataset} containing 9987 images of cats with 9 manually annotated landmarks. This set is extremly challenging due to the small number of landmarks and the high variance in head pose, optical difference between different cat races and image quality. We used 5 shape parameters and 4 global transformation parameters. 8379 images were picked randomly from the dataset and used for training and the remaining 1608 images were used for testing.
	\item The \textbf{JSRT database}~\cite{shiraishi2000development} containing 247 manually annotated chest radiographs. This dataset was selected because it is challenging due to the low contrast of radiographs and most importantly due to its small size, a common challenge in medical segmentation tasks where obtaining large numbers of images is difficult and annotations need to be generated by clinical experts. We evaluated the localization performance for lung landmarks; 209 images were used for training and 38 for testing with 25 shape parameters.
	\item The \textbf{PROMISE12} prostate dataset~\cite{litjens2014evaluation} containing 90 manually annotated prostate MRI datasets in 3D (50 for training and 40 for testing). Since the presented method works in 2D, we extracted 778 slices containing prostate segmentations from the training set and 231 slices from the test set and applied or method in 2D. 10 shape parameters were used as the shape variance of prostates is lower than in human faces.
\end{itemize}
For all tasks, the images were cropped around the landmarks with a boundary of 20\% of the landmark width and height and rescaled to 224 x 224 pixels before being fed into the network that was trained for 150 epochs using ADAM optimization. All results are shown in Fig.~\ref{fig:all_errors}.

\section{Discussion}
\label{sec:discussion}
Results indicate that the proposed method allows a super-realtime landmark detection on state-of-the-art data sets. We were able to show that our method allows efficient landmark detection under strict run-time limitations and is not limited to RGB data but can also be applied to otehr imaging modalities. Due to the diversity of tested datasets we assume that the method generalizes well towards different modalities and problems, therefore it should be applicable to other types of shape fitting problems.

\section{Conclusion}
\label{sec:conclusion}
The presented method allows extremly fast landmark detection, additionally capsuling the model generation in a fully differentiable network layer that allows its convenient use. We were able to show that our method allows landmark localization on both RGB and thermal data as well as different medical imaging modalities.

\newpage

\label{sec:ref}

\bibliographystyle{IEEEbib}
\bibliography{ICIP2019}

\end{document}